\documentclass[aps,pre,twocolumn,groupedaddress]{revtex4}
\usepackage{graphicx}
\usepackage{amsmath}

\begin{document}
\title{Do Reservoir Computers Work Best at the Edge of Chaos? }
\author{T. L. Carroll}
\email{Thomas.Carroll@nrl.navy.mil}
\affiliation{US Naval Research Lab, Washington, DC 20375}

\date{\today}

\begin{abstract}
It has been demonstrated that cellular automata had the highest computational capacity at the edge of chaos \cite{packard1988,langton1990,crutchfield1990}, the parameter at which their behavior transitioned from ordered to chaotic. This same concept has been applied to reservoir computers; a number of researchers have stated that the highest computational capacity for a reservoir computer is at the edge of chaos, although others have suggested that this rule is not universally true. Because many reservoir computers do not show chaotic behavior but merely become unstable, it is felt that a more accurate term for this instability transition is the "edge of stability"Here I find two examples where the computational capacity of a reservoir computer decreases as the edge of stability is approached; in one case, because generalized synchronization breaks down, and in the other case because the reservoir computer is a poor match to the problem being solved. The edge of stability as an optimal operating point for a reservoir computer is not in general true, although it may be true in some cases.
\end{abstract}

\maketitle

{\bf
A reservoir computer is a way of using a high dimensional dynamical system to do computation. It was demonstrated in the 1990's that dynamical systems had their greatest complexity just before they made a transition from ordered behavior to chaos (or instability), hence it was argued that the greatest computational capacity for a dynamical system should come at this "edge of chaos". Based on these concepts, it is often stated that the largest computational capacity for a reservoir computer should come at the edge of chaos; studies of reservoir computers, however, are often limited to a few node types. In many other node types, the reservoir computer never displays chaos, but becomes unstable, in which case its variables may diverge without bound. A more accurate term for the transition from ordered to disordered behavior is therefore the "edge of stability" . I show here that if many different types of node are considered, there are effects which decrease computational capacity even as the complexity of the reservoir computer dynamics increases, so the highest computational capacity is not always at the edge of stability.
}

\section{Introduction}
A reservoir computer is a high dimensional dynamical system that may be used for computation  \cite{jaeger2001,natschlaeger2002}. Reservoir computers are similar to neural networks in that they are usually built by connecting a large number of nonlinear nodes into a network, but unlike neural networks, the connections between nodes are never changed. Training of a reservoir computer takes place by creating a linear combination of the node output signals to fit a training signal. The result of the training is a set of fit coefficients. For subsequent computations, the node signals are multiplied by these coefficients to produce an output that is the result of the computation.

It has been shown that reservoir computers may do useful computations, such as reconstruction and prediction of chaotic attractors \cite{lu2018,zimmerman2018,antonik2018,lu2017,jaeger2004}, recognizing speech, handwriting or other images \cite{jalavand2018} or controlling robotic systems \cite{lukosevicius2012}.   Examples of reservoir computers so far include photonic systems \cite{larger2012, van_der_sande2017}, analog circuits \cite{schurmann2004}, mechanical systems \cite{dion2018} and  field programmable gate arrays \cite{canaday2018}. 

There are a number of criteria that must be satisfied for a reservoir computer to give optimum performance; it must produce sufficiently complex signals, but it should also be in a state of strong generalized synchronization with the driving system  \cite{lu2018, lymburn2019, hart2020, grigoryeva2020}, and the reservoir signals must be a good match for the training signal. In this paper I demonstrate reservoir computers where only some of these conditions hold.

\subsection{Edge of Chaos (Edge of Stability)}
The idea that a dynamical system or a cellular automaton has its greatest computational capacity at the edge of chaos was introduced in \cite{packard1988,langton1990,crutchfield1990}. The edge of chaos is a bifurcation between an ordered state and a disordered state. Systems in the vicinity of this phase transition exhibit the most complex behavior of any parameter range, and thus have the greatest capacity for computation. While the original work involved cellular automata, many papers on reservoir computers assume that a reservoir computer will function best at this edge. There has been some questioning of this edge of chaos principle \cite{mitchell1993, lukosevicius2007}. Many simulations have confirmed the edge of chaos rule, but the node types in many reservoir computers have been limited to nodes based on hyperbolic tangents; reservoir computers built as analog systems may have different node nonlinearities.

In many reservoir computers, depending on the node type, there is no actual chaotic behavior. Instead, when the largest Lyapunov exponent for the reservoir becomes positive, the reservoir network becomes unstable; in simulations, the reservoir signals diverge to positive or negative infinity. It is more accurate to call the point where the Lyapunov exponent becomes positive the edge of stability, so I will use that term instead of edge of chaos.

I will show below that while the entropy of a reservoir computer does increase towards the edge of stability, for some node types the best performance for the reservoir computer does not always come at this parameter value. The reservoir computer depends on generalized synchronization between the system providing the input signal and the reservoir computer network; if generalized synchronization does not exist, the reservoir computer performance will be poor \cite{lu2018, lymburn2019, hart2020, grigoryeva2020}. In some cases, generalized synchronization between the reservoir computer and the input system may exist, but the signals produced by the reservoir computer are a poor match for the task at hand, which will also lead to degraded performance. I will illustrate these points using two different reservoir computers and two different input systems. 

\section{Reservoir Computers}
The two reservoir computers were designed to be able to represent a wide variety of node types. Both reservoir computers were based on third order polynomials; one reservoir computer was an ordinary differential equation (ODE) while the other type was a map.

The polynomial ODE reservoir computer was described by:

\begin{align}
\label{res_comp}
& \frac{{d{r_i}\left( t \right)}}{{dt}}  \\
& = \alpha \left[ {{p_1}{r_i}\left( t \right) + {p_2}r_i^2\left( t \right) + {p_3}r_i^3\left( t \right) + \sum\limits_{j = 1}^M {{A_{ij}}{r_j}\left( t \right)}  + {W_i}s\left( t \right)} \right].
\end{align}

The input signal was $s(t)$ normalized to have a mean of 0 and a standard deviation of 1. The matrix ${\bf A}$ was an adjacency matrix that described how the different nodes were coupled to each other, while the vector ${\bf W}$ indicated how the input signal $s(t)$ coupled into each node. These equations were numerically integrated with a time step of 0.02. This reservoir computer model was introduced in \cite{carroll2018}.

The polynomial map reservoir computer was:
\begin{align}
\label{polymap}
&  {r _i}\left( {n + 1} \right) = \\
& \alpha \left( {{p_1}{r _i}\left( n \right) + {p_2}r _i^2\left( n \right) + {p_3}r _i^3\left( n \right) + \sum\limits_{j = 1}^M {{A_{i,j}}{r _j}\left( n \right) + {W_i}s\left( n \right)} } \right)
\end{align}
where ${\bf A}$ and ${\bf W}$ were the same as for the ODE reservoir computer.

For all simulations, the reservoir computers contained 100 nodes. The adjacency matrix was created by randomly selecting half of its entries and setting them to values drawn from a uniform random distribution between -1 and 1. The diagonal elements of ${\bf A}$ were then set to zero. In the simulations below, ${\bf A}$ was renormalized to set the spectral radius $\sigma$, the largest magnitude of its complex eigenvalues, to a specified value. The actual value of $\sigma$ was different for different simulations. The elements of ${\bf W}$ were also drawn from a uniform random distribution between -1 and 1. Because both ${\bf A}$ and ${\bf W}$ affect the performance of the reservoir computers, the structure of  ${\bf A}$ was the same for all simulations, but the overall scale of ${\bf A}$ was changed by changing the spectral radius. The same input vector ${\bf W}$ was used for all simulations. The parameters $p_1$, $p_2$, $p_3$ and $\alpha$ were chosen by random parameter searches to find reservoir computers whose best performance was at the edge of stability and other reservoir computers whose best performance was not at the edge of stability.

In the training stage, the reservoir computer is driven with the input signal $s(t)$ (or $s(n)$) to produce the reservoir computer output signals $r_i(t)$. The first 1000 points from the $r_i(t)$ time series are discarded and the next 10,000 points are used to fit a training signal $g(t)$. The actual fit signal is $h\left( t \right) = \sum\limits_{i = 1}^M {{c_i}{r_i}\left( t \right)}$, where the fit is usually done by a ridge regression to avoid overfitting. The training error is ${\Delta _{RC}} = {{\left\langle {g\left( t \right) - h\left( t \right)} \right\rangle } \mathord{\left/
 {\vphantom {{\left\langle {g\left( t \right) - h\left( t \right)} \right\rangle } {\left\langle {g\left( t \right)} \right\rangle }}} \right.
 \kern-\nulldelimiterspace} {\left\langle {g\left( t \right)} \right\rangle }}$, where $\left\langle \; \right\rangle $ indicates a standard deviation. For computation, a new input signal $s'(t)$ drives the reservoir computer. To test the computational accuracy, if the actual output that should correspond to $s'(t)$ is $g'(t)$ and the fit signal in the testing stage is $h'\left( t \right) = \sum\limits_{i = 1}^M {{c_i}{r_i}'\left( t \right)}$, then the testing error is
\begin{equation}
\label{test_err}
{\Delta _{tx}} = \frac{{\left\langle {g'\left( t \right) - h'\left( t \right)} \right\rangle }}{{\left\langle {g'\left( t \right)} \right\rangle }}
\end{equation}

\section{Input Signals}
The Lorenz system was used to generate input and training signals  \cite{lorenz1963}
\begin{equation}
\label{lorenz}
\begin{array}{l}
\frac{{dx}}{{dt}} = {c_1}y - {c_1}x\\
\frac{{dy}}{{dt}} = x\left( {{c_2} - z} \right) - y\\
\frac{{dz}}{{dt}} = xy - {c_3}z
\end{array}
\end{equation}

with $c_1$=10, $c_2$=28, and $c_3$=8/3. The equations were numerically integrated with a time step of $t_s=0.02$. The Lorenz $x$ signal was used as the input signal for all reservoir computers, while the training signal was the Lorenz $z$ signal.

A three dimensional nonlinear map  was also used to create input and training signals. The map was described by:
\begin{equation}
\label{ndmap}
\begin{array}{*{20}{l}}
{{\bf{x}}\left( {n + 1} \right) = \left[ {\begin{array}{*{20}{c}}
{1.1}&0&1\\
1&0&0\\
0&1&0
\end{array}} \right]{\bf{y}}\left( n \right)}\\
{{y_1}\left( n \right) = \,\bmod \,\left[ {{x_1}\left( n \right),1} \right]}
\end{array}
\end{equation}.

 The reservoir input signal was $x_1$ and the training signal was the $x_3$ signal. 

\section{Simulations}
For the following simulations, random parameter searches were used to find reservoir computers for which the smallest training error was at the edge of stability and other reservoir computers whose best performance was not at the edge of stability. Several statistics were computed from the reservoir computer simulations. Along with the testing error $\Delta_{tx}$ defined in eq. (\ref{test_err}), the largest Lyapunov exponent for the reservoir, its Kaplan-Yorke dimension, its entropy, and an estimate of the probability that there was a continuous function between the system providing the input signal and the reservoir were found. In addition, a spectral difference statistic defined below (eq. \ref{spectdiff}) was used.

The four largest Lyapunov exponents for the reservoir computer were calculated using a Gram-Schmidt method \cite{Parker:1989}.  To compute the entropy, each individual node time series was transformed into a symbolic time series using the ordinal pattern method \cite{bandt2002}, with a window length of four time steps. At each time step, the symbols from each node were combined into an overall symbol. The potential symbol space was huge, but because the nodes were driven by a common input signal, only a very small fraction of the symbol space was occupied. When the polynomial ODE reservoir was driven by the Lorenz $x$ signal, only 20 different symbols were observed for the entire reservoir. The entropy may then be calculated from the probability for each symbol; if $K$ total symbols have been observed, then the entropy is
\begin{equation}
\label{entropy}
H =  - \sum\limits_{k = 1}^K {p\left( {{\sigma _k}} \right)} \ln \left( {p\left( {{\sigma _k}} \right)} \right)
\end{equation}
where $p(\sigma_k)$ is the probability for symbol $\sigma_k$.

The continuity estimate was based on a statistic first described in \cite{pecora1995a} and modified in \cite{carroll2018}. We have a mapping $f$ from a space $X$ to a space $Y$, where $\left\| {} \right\|$ indicates the Euclidean metric. The function $f$ is continuous at a point ${{\bf{x}}_0} \in X$ if for every $ \varepsilon  > 0$ there exists $\delta  > 0$ such that $\left\| {{\bf{x}} - {{\bf{x}}_0}} \right\| < \delta  \Rightarrow \left\| {f\left( {\bf{x}} \right) - f\left( {{{\bf{x}}_0}} \right)} \right\| < \varepsilon $. The papers \cite{pecora1995a} and \cite{carroll2018} discuss methods to choose $\varepsilon$ and $\delta$. The result is a statistic 
\begin{equation}
\label{cont}
\psi  = [0,1]
\end{equation}
which has a range from 0 to 1, where a higher number indicates that there is a higher probability that $f$ is a continuous function.

 The Kaplan-Yorke dimension \cite{frederickson1983} is an estimate of the capacity (or fractal) dimension based on the spectrum of Lyapunov exponents of a chaotic system. For a spectrum of Lyapunov exponents $\lambda_1 \ge \lambda_2 ... \ge \lambda_d$, the Kaplan-Yorke (or Lyapunov) dimension is
\begin{equation}
\label{dimky}
{D_{KY}} = j + \sum\limits_{k = 1}^j {\frac{{{\lambda _k}}}{{\left| {{\lambda _{j + 1}}} \right|}}} 
\end{equation}
where $j$ is the largest integer for which the cumulative sum of the Lyapunov exponents is greater than 0 and the $\left| \; \right|$ operator indicates the absolute value.

The reservoir computers were simulated with many different random parameter combinations in order to find parameters for which the best performance was at the edge of stability and parameters for which the best performance was not at the edge of stability.

\subsection{Polynomial ODE Reservoir Computer}
\begin{figure}
\centering
\includegraphics[scale=0.7]{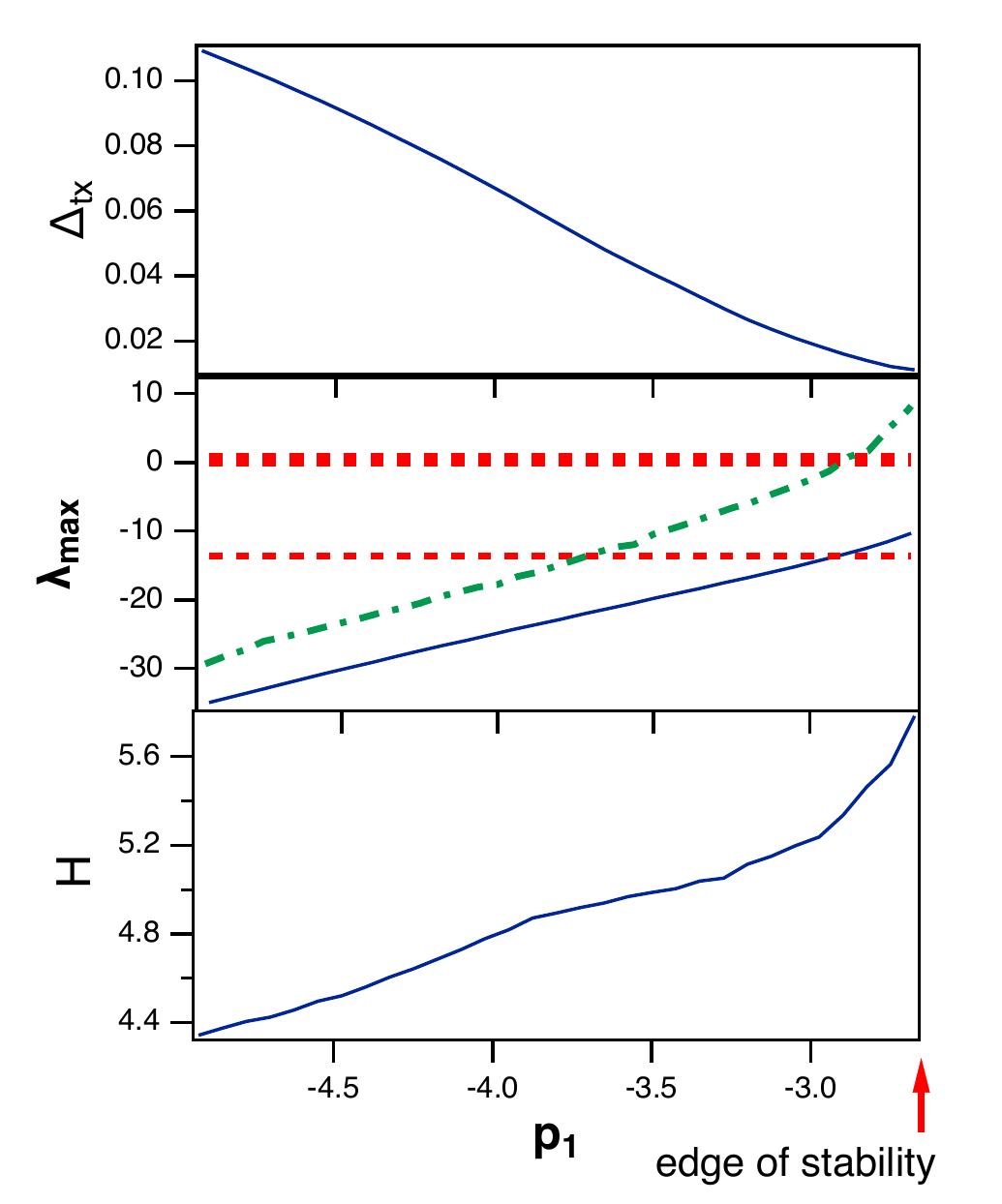} 
  \caption{ \label{nleq_lor_edge} From top to bottom, the testing error $\Delta_{tx}$, the maximum Lyapunov exponent $\lambda_{max}$ and the entropy $H$ for the polynomial ODE reservoir computer when the input signal was the Lorenz $x$ signal and the training signal was the Lorenz $z$ signal. The parameter $p_1$ was varied, while the other parameters were $p_2=-0.871984$, $p_3=0.52492$, the spectral radius $\sigma$ was 0.28512 and $\alpha$ was 5.53275. The dashed red lines in the middle graph are the Lyapunov exponents for the Lorenz system, while the green dot-dash line is the largest local Lyapunov exponent for the reservoir computer. The positive local Lyapunov exponents caused the reservoir computer became unstable for values of $p_1$ greater than those shown in the plot. When the reservoir computer became unstable, the values of the reservoir variables diverged to $\pm \infty$. }
  \end{figure}

Figure \ref{nleq_lor_edge} shows from top to bottom the testing error $\Delta_{tx}$, the maximum Lyapunov exponent for the reservoir computer ($\lambda_{max}$) and the information entropy for the reservoir computer. The parameters for this figure were $p_2=-0.871984$, $p_3=0.52492$, the spectral radius $\sigma$ was 0.28512 and $\alpha$ was 5.53275. The middle plot in figure \ref{nleq_lor_edge} also shows the maximum of the local Lyapunov exponent for the reservoir computer. The local Lyapunov exponent was calculated over a period of one time step. The positive local Lyapunov exponent caused the reservoir computer became unstable before the global Lyapunov exponent became positive. If the reservoir computer variables enter a region of large positive local Lyapunov exponent, the variables may burst to a large amplitude from which the reservoir computer does not recover, making the reservoir computer unstable even though the global Lyapunov exponent is negative.

In figure \ref{nleq_lor_edge}, the smallest testing error comes just before the reservoir computer becomes unstable as $p_1$ increases. As the reservoir computer comes closer to the edge of stability, the entropy $H$ increases. This behavior fits the expected pattern where the best computational performance comes at the edge of stability.

\begin{figure}
\centering
\includegraphics[scale=0.7]{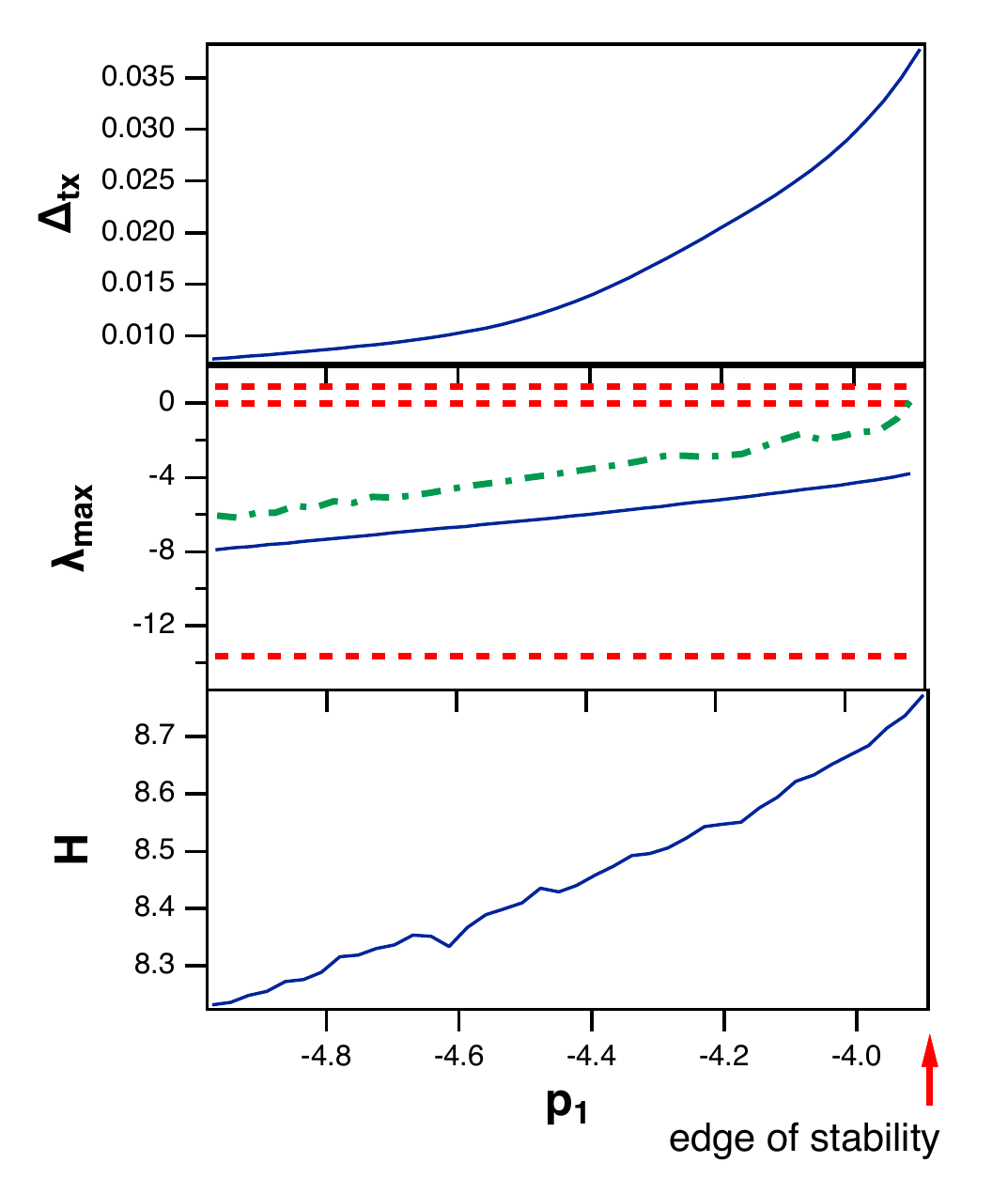} 
  \caption{ \label{nleq_lor_noedge} From top to bottom, the testing error $\Delta_{tx}$, the maximum Lyapunov exponent $\lambda_{max}$ and the entropy $H$ for the polynomial ODE reservoir computer when the input signal was the Lorenz $x$ signal and the training signal was the Lorenz $z$ signal. The parameter $p_1$ was varied, while the other parameters were $p_2=-1.03594$, $p_3=0.9308149$, the spectral radius $\sigma$ was 2.78752 and $\alpha$ was 2.72261. The dashed red lines in the middle graph are the Lyapunov exponents for the Lorenz system, while the green dot-dash line is the largest local Lyapunov exponent for the reservoir computer. The positive local Lyapunov exponents caused the reservoir computer became unstable for values of $p_1$ greater than those shown in the plot. When the reservoir computer became unstable, the values of the reservoir variables diverged to $\pm \infty$.}
  \end{figure}

Figure \ref{nleq_lor_noedge} shows a different behavior. In figure \ref{nleq_lor_noedge}, the smallest training error is not at the edge of stability. In figure \ref{nleq_lor_noedge}, the parameters were $p_2=-1.03594$, $p_3=0.9308149$, the spectral radius $\sigma$ was 2.78752 and $\alpha$ was 2.72261. Once again, positive local Lyapunov exponents (green dot-dash line) cause the reservoir computer to become unstable while the global Lyapunov exponent is still negative.

\subsubsection{Embedding Quality}
\begin{figure}
\centering
\includegraphics[scale=0.7]{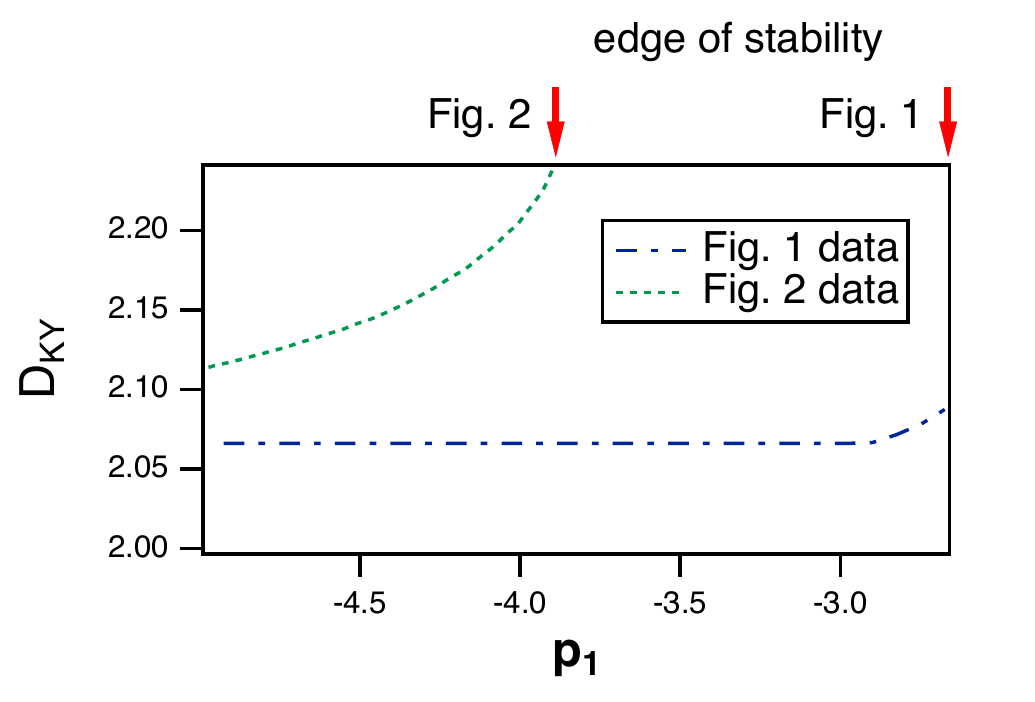} 
  \caption{ \label{nleq_lor_kydim} Kaplan-Yorke dimension $D_{KY}$ as a function of $p_1$ for the polynomial ODE reservoir computer driven by the Lorenz $x$ signal for the parameter configurations in figures \ref{nleq_lor_edge} and \ref{nleq_lor_edge}. "Fig. 1" refers to the parameters in figure \ref{nleq_lor_edge}, where the best performance came at the edge of stability, while "Fig. 2" refers to the parameters in figure \ref{nleq_lor_noedge}, where the best performance was not at the edge of stability.}
  \end{figure}

In figure \ref{nleq_lor_noedge}, the maximum Lyapunov exponent for the reservoir computer lies in between the Lyapunov exponents for the Lorenz system, leading to an increase in the Kaplan-Yorke dimension for the reservoir computer. The Kaplan-Yorke dimensions for the parameter configurations in both figures \ref{nleq_lor_edge} and \ref{nleq_lor_noedge} are shown in figure \ref{nleq_lor_kydim}.

The Kaplan-Yorke dimension for the Lorenz system used in this work is 2.06. Figure \ref{nleq_lor_kydim} shows that when the smallest testing error comes at the edge of stability, there is only a very small increase in the Kaplan-Yorke dimension of the reservoir signals, while the Kaplan-Yorke dimension shows a larger increase when the smallest testing error does not come at the edge of stability. Because the dimension of the reservoir computer signals differs from the dimension of the Lorenz system, the reservoir computer is not an embedding of the driving system, leading to a degradation in performance. Examples of how the continuity and differentiability between a driven dynamical system and the driving system vary as parameters are changed are shown in \cite{pecora2000}. 

Figure \ref{nleq_lor_kydim} does show a small increase in Kaplan-Yorke dimension just before the edge of stability for the data from fig. \ref{nleq_lor_edge}, in which the smallest testing error is at the edge of stability. Because this increase is small, it is presumably offset by the increase in entropy near the edge of stability, also seen in fig. \ref{nleq_lor_edge}. According to \cite{packard1988,langton1990,crutchfield1990}, increased entropy should lead to increased computational capacity, and the dimension change here is not large enough to overcome the entropy increase.

\begin{figure}
\centering
\includegraphics[scale=0.7]{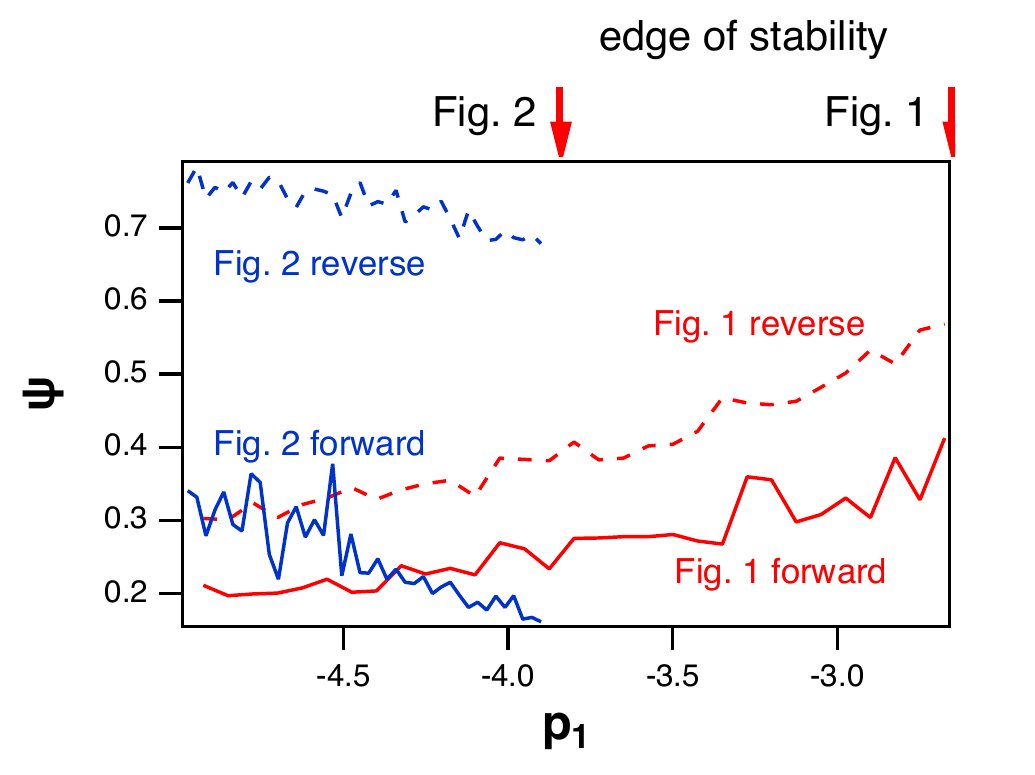} 
  \caption{ \label{nleq_lor_cont} Foward and reverse continuity statistics $\psi$ as a function of $p_1$ for the polynomial ODE reservoir computer driven by the Lorenz $x$ signal for the parameter configurations in figures \ref{nleq_lor_edge} and \ref{nleq_lor_edge}.  "Fig. 1" refers to the parameters in figure \ref{nleq_lor_edge}, where the best performance came at the edge of stability, while "Fig. 2" refers to the parameters in figure \ref{nleq_lor_noedge}, where the best performance was not at the edge of stability. For the forward statistic, the $\delta$ neighborhood is on the full Lorenz system while the $\epsilon$ neighborhood is on the reservoir computer. For this statistic, 1 means a high probability that there is a continuous function and 0 means a low probability.}
  \end{figure}

The continuity statistic for the polynomial ODE reservoir computer echos the Kaplan-Yorke dimension statistic. Figure \ref{nleq_lor_cont} shows both forward and reverse continuity statistics. For the forward statistic, the $\delta$ neighborhood is on the full Lorenz system while the $\epsilon$ neighborhood is on the reservoir computer.

When the parameters for the reservoir computer are such that the smallest testing error is at the edge of stability, the continuity statistic increases as $p_1$ increases, indicating an increased probability of a continuous function both from the full Lorenz system to the reservoir computer and from the reservoir computer to the full reservoir system. Figure \ref{nleq_lor_edge} showed that the testing error also decreased as $p_1$ increased.

When the parameters for reservoir computer are set so that the smallest testing error is not at the edge of stability, both forward and reverse continuity decrease as $p_1$ increases. In this configuration, there is a lower probability of a continuous function between the Lorenz system and the reservoir, and the testing error increases, as $p_1$ increases. The lower continuity is caused by the change in the dimension of the reservoir computer signals. This change in dimension is also known as a transition between weak and strong synchronization of chaos \cite{pyragas1996}.

To summarize, when the polynomial ODE reservoir computer was driven with the Lorenz $x$ signal, if the Lyapunov exponent spectrum of the reservoir computer overlaps with the Lyapunov spectrum of the Lorenz system by a sufficient amount, the Kaplan-Yorke dimension of the reservoir computer can increase, so that the reservoir computer is no longer in a state of generalized synchronization with the Lorenz system, resulting in an increase in the training and testing errors.

\subsection{Polynomial Map Reservoir Computer with Map Input}

\begin{figure}
\centering
\includegraphics[scale=0.7]{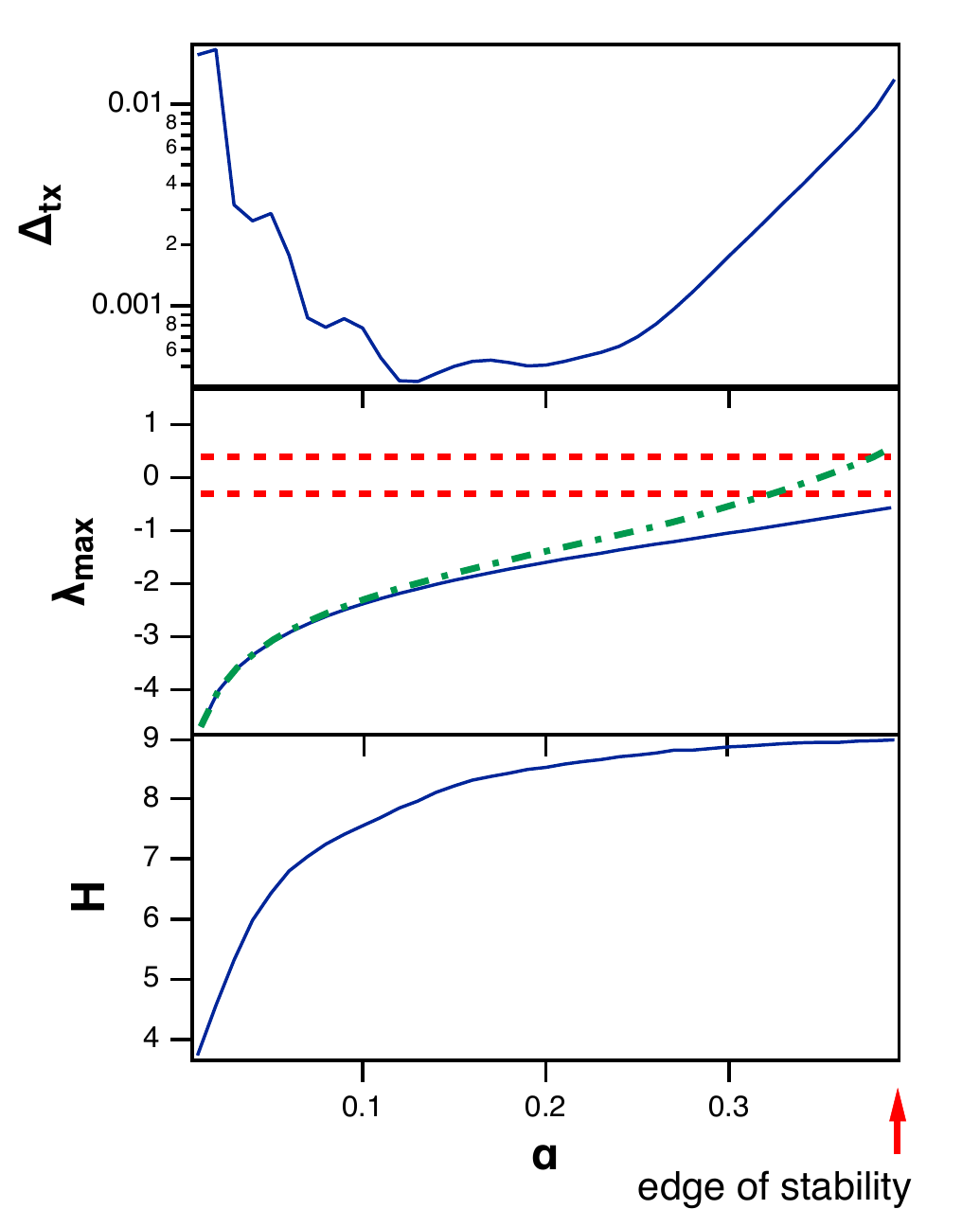} 
  \caption{ \label{pmap_map_noedge} From top to bottom, the testing error $\Delta_{tx}$, the maximum Lyapunov exponent $\lambda_{max}$ and the entropy $H$ for the polynomial map reservoir computer when the input signal was the $x_1$ signal from the 3d map of eq. (\ref{ndmap}) and the training signal was the corresponding $x_3$ signal. The parameter $\alpha$ was varied, while the other parameters were  $p_1= 0.5$, $p_2=0.5$, $p_3=0.5$, and the spectral radius $\sigma$ was 0.5. The red dashed lines in the middle graph are the Lyapunov exponents for the 3d map system, while the green dot-dash line is the largest local Lyapunov exponent for the reservoir computer. The positive local Lyapunov exponents caused the reservoir computer to become unstable for values of $\alpha$ larger than those shown in the plots. When the reservoir computer became unstable, the values of the reservoir variables diverged to $\pm \infty$.}
  \end{figure}

The parameter chosen for the single parameter sweep in the polynomial map reservoir computer was the time scale parameter $\alpha$. Figure \ref{pmap_map_noedge} shows from top to bottom the testing error $\Delta_{tx}$, the maximum Lyapunov exponent for the reservoir computer ($\lambda_{max}$) and the entropy for the polynomial map reservoir computer when the input signal was the 3d map $x_1$ signal and the training signal was the $x_3$ signal.  The parameters for this figure were $p_1= 0.5$, $p_2=0.5$, $p_3=0.5$, and the spectral radius $\sigma$ was 0.5. These parameters were chosen to demonstrate a situation in which the optimal reservoir performance was not at the edge of stability. The middle plot in figure \ref{pmap_map_noedge} also shows the maximum local Lyapunov exponent for the reservoir as a green dot-dash line. The positive local Lyapunov exponents cause the reservoir computer to become unstable while the global Lyapunov exponents are still negative. 

\begin{figure}
\centering
\includegraphics[scale=0.7]{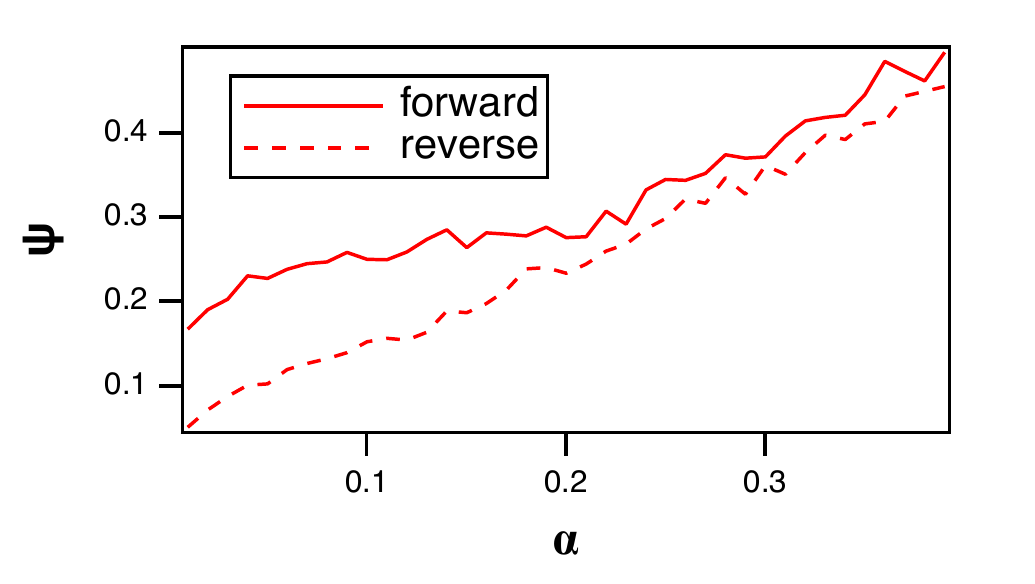} 
  \caption{ \label{pmap_cont_noedge} Forward and reverse continuity $\psi$ as a function of the time scale parameter $\alpha$ for the polynomial map reservoir computer driven by the $x_1$ signal from the 3d map.
}
\end{figure}

The plots for the polynomial map reservoir computer driven by the 3d map signal in figure \ref{pmap_map_noedge} do not fit the same pattern as the simulations for the polynomial ODE. The Lyapunov exponent spectrum of the polynomial map reservoir computer does not overlap with the Lyapunov exponent spectrum of the 3d map system, but the smallest testing error is still not at the edge of stability. Figure \ref{pmap_cont_noedge} further illustrates this contrast.

As the parameter $\alpha$ in figure \ref{pmap_cont_noedge} approaches the edge of stability, the forward and reverse continuity both increase. This increase indicates that the probability that there is a continuous function from the full 3d map to the reservoir computer and also in the reverse direction increases towards the edge of stability. The presence of a continuous function indicates generalized synchronization, and should lead to good reservoir computer performance.

\begin{figure}
\centering
\includegraphics[scale=0.7]{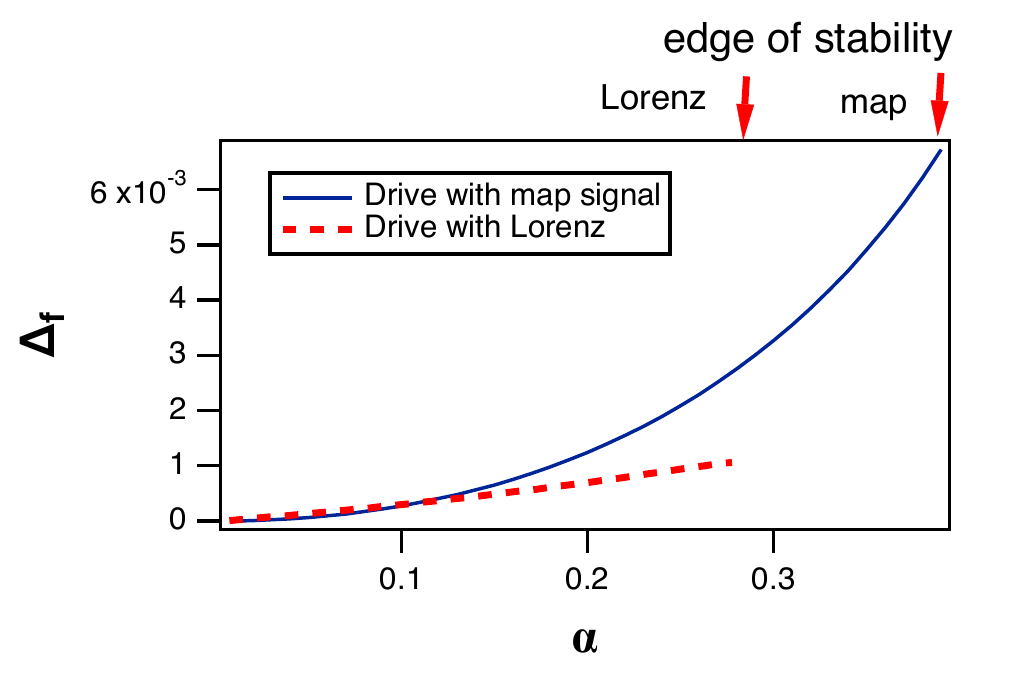} 
  \caption{ \label{sdiff_noedge} Spectral difference statistic $\Delta_f$ as defined in eq. (\ref{spectdiff}) as a function of $\alpha$. The solid blue line is for the polynomial map reservoir computer driven by the $x_1$ signal from the 3d map, while the red dashed line is for the same reservoir computer driven by the Lorenz $x$ signal, as described in section \ref{pmap_lor}. Note that the location of the edge of stability depends on the input signal.
}
\end{figure}

One factor that has not been considered in the statistics presented so far is how well the reservoir signals match the training or testing signals. In order to quantify the match between the Fourier spectra of the training signals and the reservoir signals, a spectral difference statistic was developed. 
Designating the Fourier transform of the reservoir computer signal for node $i$ as $R_i(f)$ and for the training signal as $G(f)$, the total power in the training signal, weighted by frequency, is $\sum\limits_{j = 1}^{{N_f}} {\left| {G\left( {{f_j}} \right)} \right|{f_j}} $, and for the time series from the $i$'th node it is $\sum\limits_{j = 1}^{{N_f}} {\left| {{R_i}\left( {{f_j}} \right)} \right|{f_j}}$, where $N_f$ is the number of discrete frequencies in the power spectrum, and $\left| \; \right|$ indicates the magnitude of the complex frequency. The weighted power will be larger if the power spectrum has a larger magnitude at higher frequencies, but it will also be larger if the signal contains more power, so these weighted powers must be normalized by the total power in the spectrum, so the normalized weighted powers are ${{\sum\limits_{j = 1}^{{N_f}} {\left| {G\left( {{f_j}} \right)} \right|{f_j}} } \mathord{\left/
 {\vphantom {{\sum\limits_{j = 1}^{{N_f}} {\left| {G\left( {{f_j}} \right)} \right|{f_j}} } {\sum\limits_{j = 1}^{{N_f}} {\left| {G\left( {{f_j}} \right)} \right|} }}} \right.
 \kern-\nulldelimiterspace} {\sum\limits_{j = 1}^{{N_f}} {\left| {G\left( {{f_j}} \right)} \right|} }}$ and ${{\sum\limits_{j = 1}^{{N_f}} {\left| {{R_i}\left( {{f_j}} \right)} \right|{f_j}} } \mathord{\left/
 {\vphantom {{\sum\limits_{j = 1}^{{N_f}} {\left| {{R_i}\left( {{f_j}} \right)} \right|{f_j}} } {\sum\limits_{j = 1}^{{N_f}} {\left| {{R_i}\left( {{f_j}} \right)} \right|} }}} \right.
 \kern-\nulldelimiterspace} {\sum\limits_{j = 1}^{{N_f}} {\left| {{R_i}\left( {{f_j}} \right)} \right|} }}$. The weighted spectral difference between the training signal and the reservoir computer is\
\begin{equation}
\label{spectdiff}
{\Delta _f} = \frac{{\sum\limits_{j = 1}^{{N_f}} {\left| {G\left( {{f_j}} \right)} \right|{f_j}} }}{{\sum\limits_{j = 1}^{{N_f}} {\left| {G\left( {{f_j}} \right)} \right|} }} - \frac{1}{M}\sum\limits_{i = 1}^M {\frac{{\sum\limits_{j = 1}^{{N_f}} {\left[ {\left( {\left| {G\left( {{f_j}} \right)} \right| - \left| {{R_i}\left( {{f_j}} \right)} \right|} \right){f_j}} \right]} }}{{\sum\limits_{j = 1}^{{N_f}} {\left( {\left| {G\left( {{f_j}} \right)} \right| - \left| {{R_i}\left( {{f_j}} \right)} \right|} \right)} }}},
\end{equation}
 The frequencies in eq. (\ref{spectdiff}) were normalized to go from 0 to 0.5. The spectral difference $\Delta_f$ shows how the weighted frequency spectrum of the reservoir computer differs from the weighted frequency spectrum of the training signal. Positive values of $\Delta_f$ mean that the training signal spectrum contains more high frequencies than the reservoir computer frequency spectrum. A plot of $\Delta_f$ as a function of $\alpha$ is shown in figure \ref{sdiff_noedge}.

The plot of $\Delta_f$ in figure \ref{sdiff_noedge} shows that as $\alpha$ increases, the reservoir computer signals contain less high frequency content than the training signal $g(n)$. As the largest Lyapunov exponent of the reservoir computer approaches zero from below, the frequency response of the reservoir computer slows- an equivalent would be to say that the memory becomes longer. For higher values of $\alpha$, the reservoir computer does not keep up with changes in the 3d map signal, so the testing error becomes larger. Another way of stating this is that as the memory of the reservoir computer becomes longer, its performance becomes worse, which is the opposite of the standard assumption.

\subsection{Polynomial Map Reservoir Computer with Lorenz Input}
\label{pmap_lor}
\begin{figure}
\centering
\includegraphics[scale=0.7]{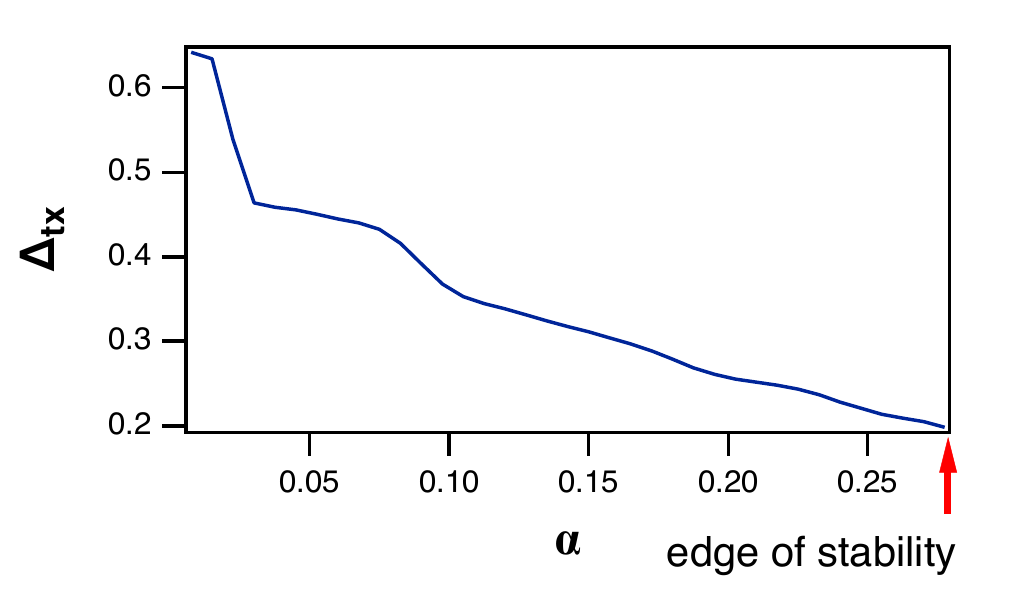} 
  \caption{ \label{pmap_lor_edge} Testing error $\Delta_{tx}$ for the polynomial ODE map as a function of $\alpha$ when the input signal is the Lorenz $x$ signal and the training signal is the Lorenz $z$ signal. The edge of stability when this reservoir computer is driven with the Lorenz $x$ signal is at a lower value of $\alpha$ then when the driving signal came from the 3d map.}
\end{figure}

As a contrast, the smallest testing error for the polynomial map reservoir computer is at the edge of stability when the input signal is the Lorenz $x$ signal and the training signal is the Lorenz $z$ signal. Figure \ref{pmap_lor_edge} shows the testing error for this combination.

Figure \ref{sdiff_noedge} shows that the spectral difference statistic for the polynomial map reservoir computer driven by the Lorenz $x$ signal does increase, meaning that the reservoir signals have less high frequency content as $\alpha$ increases, but the spectrum difference from the training signal is not as great as when the reservoir computer was driven by the 3d map signal. In this case, the loss of high frequencies is compensated by the increase in entropy of the reservoir signals, so the testing error continues to decrease up to the edge of stability.

Figures \ref{pmap_map_noedge} and \ref{pmap_lor_edge} show that the same reservoir computer can yield the smallest testing error at the edge of stability or away from the edge of stability, depending on the problem being solved. In this section, generalized synchronization between the input system and the reservoir computer was maintained, but a poor match between the reservoir signals and the Lorenz signals causes large testing and training signals.

\section{Summary}
The concept that the best computational capacity for a dynamical system comes at the edge of stability is not true in general, and this paper shows reservoir computers where the best performance is not at the edge of stability. For the reservoir computers simulated here, the maximum entropy did come at the edge of stability, but there were other dynamical effects that caused the testing and training errors to become larger as the reservoir computer approached this edge. Overlap in the Lyapunov exponent spectra of the driving system and the reservoir computer could cause a transition from strong to weak generalized synchronization, or changes in the frequency response of the reservoir computer could result in a poor match between the reservoir computer signals and the training signal.

The simulations here and the work in \cite{lu2018, lymburn2019, hart2020, grigoryeva2020} claim that reservoir computers should be in a state of generalized synchronization to the input system for tasks such as signal fitting and prediction. They did not study classification tasks; whether generalized synchronization is necessary for classification is an open question.

 \section{Data Availability}
The data that support the findings of this study are available on request from the corresponding author. The data are not publicly available because they have not been approved for public release.

This work was supported by the Naval Research Laboratory's Basic Research Program.

\end{document}